\documentclass[letterpaper]{article} 
\usepackage[preprint]{aaai2027}  
\usepackage[hyphens]{url}  
\usepackage{graphicx} 
\usepackage{natbib}  
\usepackage{caption} 
\AddToHook{env/equation/begin}{\small}
\AddToHook{env/displaymath/begin}{\small}
\usepackage{amsmath}
\usepackage{amssymb}
\usepackage{amsfonts}
\usepackage{booktabs}
\usepackage{multirow}
\usepackage{xcolor}
\usepackage{tikz}
\usetikzlibrary{arrows.meta,positioning}
\usepackage{algorithm}
\usepackage{algorithmic}
\usepackage[most]{tcolorbox}
\usepackage{color}
\usepackage{soul}

\newcommand{\piS}{\pi_{S}}
\newcommand{\piT}{\pi_{T}}
\newcommand{\PS}{P_t^{S}}
\newcommand{\PT}{P_t^{T}}

\newcommand{\Ls}{\mathcal{L}_{S}}
\newcommand{\Lt}{\mathcal{L}_{T}}
\newcommand{\Ltot}{\mathcal{L}_{\mathrm{total}}}
\newcommand{\Ex}{\mathbb{E}}

\tcbset{
  vistaPromptBox/.style={
    enhanced,
    colback=white,
    colframe=black!75,
    colbacktitle=white,
    coltitle=black,
    boxrule=0.55pt,
    arc=2pt,
    left=5pt,
    right=5pt,
    top=4pt,
    bottom=4pt,
    fonttitle=\bfseries,
    fontupper=\small,
    title style={draw=none,fill=white},
    titlerule style={draw=black!75,line width=0.55pt},
    before skip=5pt,
    after skip=6pt
  }
}

\title{VISTA: Verifier-Informed Student-to-Teacher Adaptation for\\
On-Policy Self-Distillation}
\author{Zewen Ding\textsuperscript{1,2}\equalcontrib, Zezhong Wu\textsuperscript{1,2}\equalcontrib, Zhou Tao\textsuperscript{1,2}, Shida Wang\textsuperscript{1,2}, Shizhuo Hou\textsuperscript{1,2}, YongXiang Hua\textsuperscript{1,2}, Haoyu Cao\textsuperscript{1,2}, Linli Xu\textsuperscript{1,2}\corresponding}
\affiliations{
\textsuperscript{1}University of Science and Technology of China\\
\textsuperscript{2}State Key Laboratory of Cognitive Intelligence
}

\begin{document}
\maketitle
\vspace{-30pt}

\begin{abstract}
On-policy self-distillation (OPSD) improves reasoning by training a problem-only
student on its own rollouts using dense token-level supervision from a
privileged teacher that also sees a reference solution. However, standard OPSD
treats the teacher distribution as a fixed target along the
student's rollout and updates only the student 
-- even though privileged conditioning
does not guarantee that the teacher always provides the most appropriate target
for problem-only reasoning. This one-way supervision can therefore misdirect the student when the teacher distribution is misaligned with valid student reasoning. We therefore
introduce Verifier-Informed Student-to-Teacher Adaptation (VISTA), which
preserves the standard OPSD student update while using outcome-verified rollouts
to adapt the teacher toward the student distribution. Within each verified
rollout, VISTA further restricts this adaptation to the top-$k$ positions with
the largest teacher--student KL divergence. Notably, VISTA
reuses the rollout and loss function from standard OPSD, introducing no
additional sampling or separate reward objective.
Across AIME24, AIME25, and HMMT25 with Qwen3 models at 1.7B, 4B, and 8B, VISTA
achieves the highest Avg@12 at every scale, improving over OPSD by $0.6$, $0.7$,
and $2.1$ points, respectively. These results demonstrate the value of student
supervision from outcome-verified rollouts and highlight student-to-teacher
adaptation as a promising direction for OPSD.

\end{abstract}

\suppressfloats[t]
\begin{figure}[t]
  \centering
  \includegraphics[width=0.9\columnwidth]{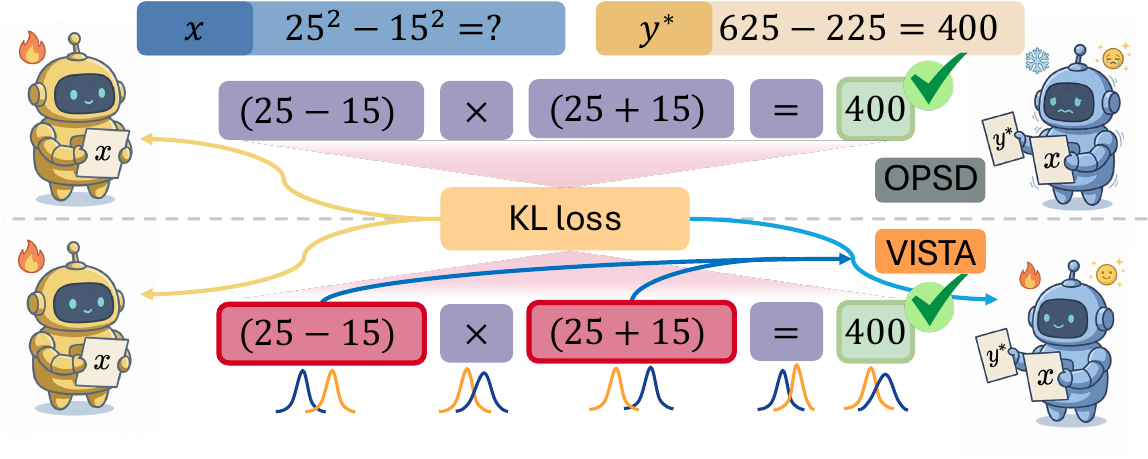}
  \caption{A schematic of VISTA. Building on on-policy self-distillation
  (OPSD), VISTA additionally introduces a selective student-to-teacher
  adaptation on verified trajectories: at token positions selected by their KL
  divergence, the privileged teacher is revised toward the student's
  distribution, so that the teacher in turn provides more benign supervision to
  the student.}
  \label{fig:teaser}
\end{figure}

\section{Introduction}
\label{sec:intro}

On-policy self-distillation (OPSD) has recently emerged as a practical way to
post-train reasoning models~\cite{opsd2026,hubotter2026sdpo}. In each OPSD step, a problem-only student samples 
a rollout using only
the information available at inference. A privileged teacher with additional access to a
reference solution then provides dense token-level supervision. These on-policy
rollouts keep training aligned with inference-time behavior, while the token-level
targets provide fine-grained guidance throughout the reasoning process.

Although this design is appealing in practice, we identify a structural
limitation of standard OPSD, which we term the
\emph{teacher-superiority assumption}: at every position along
the student's rollout, standard OPSD treats the privileged distribution as the
better token-level target, even though reference conditioning alone does not
guarantee its superiority for problem-only reasoning.
This assumption can misdirect the student in two ways. First, the teacher may
favor tokens that explicitly invoke its privileged input, such as ``reference
solution.'' Training toward such targets 
can lead the problem-only student to
mention a reference it never observes at inference
~\cite{yang2026rlsd,shen2026purified}. Second, when a valid token
favored by the student departs from the teacher's preferred continuation, the
teacher may assign it too little probability. Using the teacher distribution as
the target lowers the student's probability for that token, suppressing valid
reasoning and self-correction
~\cite{kim2026rebellious,shen2026antisd,
kaur2026rethinking,peng2026adopsd}.

Recent methods reduce these failures on the student side. RLSD anchors update
directions to verifier rewards rather than the teacher distribution
~\cite{yang2026rlsd}, whereas TRACE confines distillation to selected token
spans~\cite{wang2026trace}, and DemoPSD reweights the student's distillation
target toward the student where teacher and student disagree~\cite{li2026demopsd}. These safeguards reduce the student's
exposure to mismatched targets, but 
supervision remains one-way from teacher to
student, leaving valuable signals in student rollouts and their token-level
distributions unused. 
This raises the following question: under appropriate conditions, can
feedback from the problem-only student's distribution improve the privileged
teacher, so that the teacher, in turn, supervises the student better? 

Letting student behavior revise the privileged teacher is not a matter of
simply reversing the distillation direction: the student’s token-level distributions are not in general better targets than the teacher’s, and adopting them wholesale could pull the teacher distribution in the wrong direction. Even
when the student can be trusted, a second problem remains: teacher--student disagreement
is
highly non-uniform across positions, and updating
the teacher at every token
could rapidly
shrink the distributional gap and erode complementary privileged knowledge.
The central design question is therefore twofold: which rollouts can provide teacher
feedback, and which token positions within them should carry it.

We introduce \emph{Verifier-Informed Student-to-Teacher Adaptation (VISTA)}, an effective framework that enables an OPSD teacher to learn from its
student. To the best of our knowledge, VISTA is the first to establish
student-to-teacher feedback within OPSD while preserving the standard student
update. Its simplicity is deliberate: VISTA answers the two questions
above through rollout-level and token-level selection. Teacher feedback is enabled only for rollouts 
that pass the
outcome verifier, since 
the verified outcome is a 
inexpensive yet reliable signal that the
student's distributions along that rollout are informative and worth learning from. Within 
each eligible rollout, KL-guided position selection adapts the teacher toward
the student at the top-$k$ positions of largest teacher--student
disagreement. Concentrating adaptation on these positions targets the
teacher distributions most misaligned with problem-only reasoning, while the
sparse selection keeps the teacher from converging on the student too
quickly and eroding its subsequent supervision. Crucially, the
method reuses the same rollouts and loss function as standard OPSD,
requiring neither additional sampling nor a separate reward objective.

When compared against 
Base, SFT, GRPO, SDPO and standard OPSD results under the
matched OPSD protocol, VISTA achieves the highest average scores (Avg@12 across three benchmarks) 
at all
three Qwen3 scales. Relative to reported OPSD, VISTA improves the
three-benchmark average by $0.6$, $0.7$, and $2.1$ points at 1.7B, 4B, and 8B scales,
respectively, and achieves state-of-the-art performance in eight of the nine scale-benchmark settings. 
Ablations 
validate the contributions of
verifier-gated and selective teacher adaptation, while three analytical studies
indicate that the adapted teacher better supports validated student reasoning,
reduces explicit reference attribution, and preserves teacher capability.

Our contributions are threefold:
\begin{itemize}\itemsep2pt
  \item We formalize the 
  limitations 
  of OPSD's teacher-superiority assumption and
  characterize how reference-conditioned distributions can over-support
  reference-specific next tokens or under-support valid alternatives.
  \item We explore whether an OPSD teacher can learn from its student and
  introduce VISTA, which enables student-to-teacher adaptation through a
  verifier-based outcome gate and a top-$k$ token-position mask while
  preserving standard OPSD student training.
  \item Across three Qwen3 scales and three competition-math benchmarks,
  VISTA achieves the highest aggregate Avg@12 at every scale and
  state-of-the-art performance in eight of the nine scale--benchmark settings
  under the matched OPSD protocol.
\end{itemize}

\begin{figure*}[t]
  \centering
  \includegraphics[width=\textwidth]{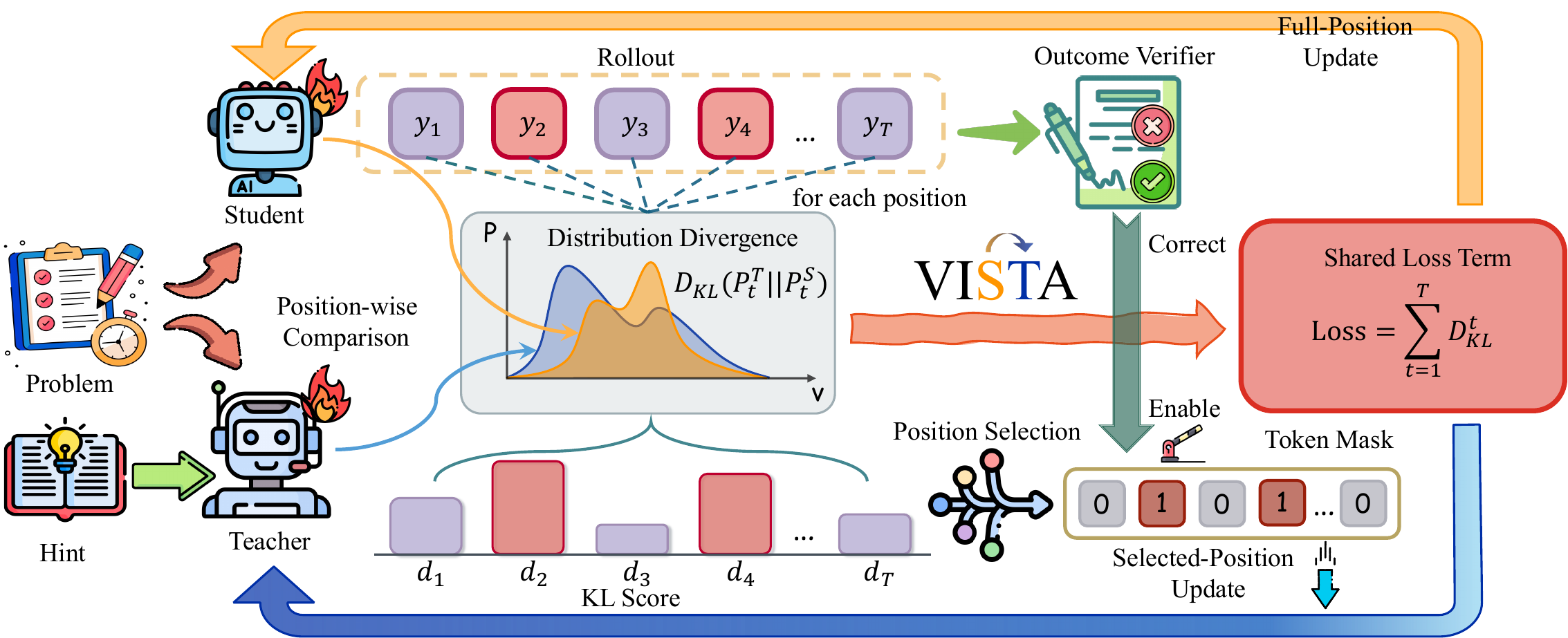}
  \caption{Overview of the VISTA training pipeline. (1)
  On-policy rollout: A problem-only student samples a solution rollout for a
  training problem. (2) Outcome gating and position selection: The
  outcome verifier gates teacher updates based on rollout correctness, while
  position selection forms the teacher-update mask by selecting the top-$k$
  positions with the largest teacher--student KL divergence. (3)
  Student and teacher updates: Building on standard OPSD, VISTA retains the
  standard student update at every rollout position and, as explained in (2),
  updates the teacher toward the student's distribution at token positions
  where the teacher-update mask equals one.}
  \label{fig:main}
\end{figure*}

\section{Related Work}
\label{sec:related}

\paragraph{On-policy distillation with privileged context.}
Knowledge distillation transfers a teacher's predictive distribution to a
student~\cite{hinton2015distilling,kim2016sequence,sanh2019distilbert,gou2021knowledge}.
Two extensions define our setting. First, \emph{privileged} distillation lets
the teacher exploit side information available only at training time;
generalized distillation and context distillation cast this privileged
information as a teacher--student
transfer~\cite{vapnik2009new,lopez2015unifying,snell2022learning}.
Second, \emph{on-policy} distillation supervises the student on its own
continuations rather than a fixed
corpus~\cite{agarwal2024gkd,gu2024minillm,ross2011dagger,xu2025speculative}.
On-policy self-distillation
(OPSD) unites both: a problem-only student trains on its own rollouts under
dense token-level targets from a teacher privileged with a reference
solution~\cite{opsd2026}. Closely related methods share this regime: OPCD
conditions the teacher on additional context~\cite{ye2026opcd}, SDPO supplies
feedback or a successful sibling attempt~\cite{hubotter2026sdpo}, and Privileged
Information Distillation applies reward objectives to teacher
trajectories~\cite{penaloza2026privileged}. All of these methods retain a fixed
privileged-to-unprivileged teaching direction. Although mutual and
self-distillation relax this direction, they do so only in plain classification,
without privileged information or outcome
verification~\cite{furlanello2018born,zhang2018deepmutual,anil2018online,zhang2019byot}. VISTA
instead lets outcome-verified student behavior revise the privileged target.

\paragraph{Verifier-informed reasoning supervision.}
In mathematics and code, outcome verifiers determine whether a rollout reaches
a valid final result and provide a reliable sequence-level
signal~\cite{cobbe2021training,uesato2022solving,le2022coderl}. RLVR turns this
verifier signal directly into a policy-gradient reward, making outcome
verification itself the learning
signal~\cite{shao2024deepseekmath,deepseekr1,lambert2024tulu3}. Different from
process supervision, which needs step-level labels or learned reward
models~\cite{uesato2022solving,lightman2023lets,wang2024mathshepherd}, this outcome check is cheap to
obtain and requires no additional annotation. Self-training and filtering
methods such as STaR, RFT, and Reinforce-Rej instead reuse successful
completions for supervised updates or outcome-filtered policy
optimization~\cite{zelikman2022star,yuan2023scaling,xiong2025minimalist}.
Verification in these methods ultimately governs updates to the model being
optimized. In VISTA, the outcome signal instead gates the teacher-side
distributional update: only
verified rollouts provide soft student targets for teacher adaptation, while the
dense OPSD loss remains active on every rollout.

\paragraph{Selective and adaptive privileged supervision.}
Dense or sustained OPSD can amplify artifacts or degrade continual
post-training~\cite{wang2026denser,shen2026purified,kaur2026rethinking,peng2026adopsd},
motivating regulation by update direction
or reasoning span: RLSD signs and scales token-level credit by the outcome
advantage~\cite{yang2026rlsd}, and TRACE routes KL supervision to
annotator-identified reasoning spans~\cite{wang2026trace}, while DemoPSD
down-weights the teacher target where the student and teacher
disagree~\cite{li2026demopsd}. These apply this
selectivity only to the \emph{student} update, leaving the privileged teacher a
fixed target. We instead make the teacher itself adaptive: outcome-verified
problem-only rollouts supply the student-distribution targets, and
full-distribution disagreement selects the update positions.

\section{Method: VISTA}
\label{sec:method}

\subsection{Preliminary}
\label{sec:prelim}
\label{sec:prelim-opsd}
On-policy self-distillation (OPSD)~\cite{opsd2026} trains a problem-only
student using next-token distributions from a privileged teacher. Let
$\mathcal{S}=\{(x_i,y_i^\star)\}_{i=1}^N$ be a training set of problems and
reference solutions. The student policy
$\piS(\cdot\mid x)$ sees only the problem $x$, while the privileged teacher
$\piT(\cdot\mid x,y^\star)$ also sees the reference solution $y^\star$.

Training remains on-policy: a rollout $y$ is sampled from the current student,
$y\sim\piS(\cdot\mid x)$, and both policies are evaluated on the same partial
student rollout. For position $t$ with prefix $y_{<t}$, define
\[
  \PS = \piS(\cdot\mid x,y_{<t}), \qquad
  \PT = \piT(\cdot\mid x,y^\star,y_{<t}).
\]
OPSD trains the student to match the privileged teacher along these sampled
rollouts:
\begin{equation}
  \mathcal{L}_{\mathrm{OPSD}}
  =
  \Ex_{\substack{(x,y^\star)\sim\mathcal{S}\\
                  y\sim\piS(\cdot\mid x)}}
  \left[
    \frac{1}{|y|}\sum_{t=1}^{|y|}
    D^{\mathrm{clip}}_{\mathcal V,\tau}\!\left(\PT\Vert\PS\right)
  \right],
  \label{eq:opsd}
\end{equation}
where $D^{\mathrm{clip}}_{\mathcal V,\tau}$ is the OPSD pointwise-clipped
full-vocabulary forward KL~\cite{kullback1951information},
\begin{equation}
  D^{\mathrm{clip}}_{\mathcal V,\tau}(P\Vert Q)
  =
  \sum_{v\in\mathcal V}
  \min\!\left(P(v)\log\frac{P(v)}{Q(v)},\tau\right).
  \label{eq:opsd-clipped-kl}
\end{equation}
Thus, OPSD applies one-way teacher-to-student supervision at every prefix of the
sampled rollout. It thereby combines on-policy training with dense token-level
feedback, without an external teacher or reward model.

\subsection{Problem: The Teacher-Superiority Assumption in OPSD}
\label{sec:feedback-gap}
At each prefix along a student-generated rollout, standard OPSD trains the student to match the teacher’s next-token distribution. However, we argue that this one-way update relies on what we call the \emph{teacher-superiority assumption}: the teacher must remain a better learning target than the current student throughout the rollout.

To formalize this assumption, let $P_t^I$ denote the unobserved next-token
distribution of an ideal problem-only policy at prefix $h_t=(x,y_{<t})$. OPSD
aims to approach this ideal distribution indirectly by moving $P_t^S$ toward
$P_t^T$. We define their relative distance to the ideal target as
\begin{equation}
  \Delta_t
  =D_{\mathrm{KL}}(P_t^I\Vert P_t^T)
   -D_{\mathrm{KL}}(P_t^I\Vert P_t^S),
  \label{eq:teacher-superiority-gap}
\end{equation}
where $\Delta_t<0$ means that the teacher is closer to $P_t^I$; for
$\Delta_t>0$, the reverse holds.

Moving $P_t^S$ toward $P_t^T$ can bring it closer to $P_t^I$ only when the teacher is itself closer to the ideal distribution than the current student. 
By Eq.~\eqref{eq:teacher-superiority-gap}, this requires $\Delta_t<0$ at every
token position, which is precisely the teacher-superiority assumption.

\begin{figure}[hbt]
  \centering
  \includegraphics[width=\linewidth]{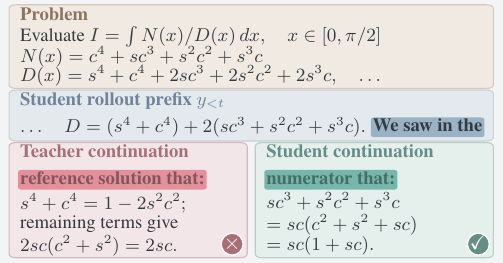}
  \caption{A representative case of the teacher-superiority assumption failing.
  From the same student rollout prefix, the student
  follows valid reasoning, whereas the teacher emits a
  reference-specific phrase. This counterexample shows
  that the privileged distribution is not uniformly a better token-level target
  along the student's rollout.}
  \label{fig:teacher-superiority-case}
\end{figure}

Reference conditioning changes the teacher’s information set, but by itself cannot guarantee that the resulting distribution is uniformly better suited to problem-only deployment. At positions where
$\Delta_t>0$, OPSD still applies the same update even though the student is
closer to $P_t^I$. This creates two risks. For a candidate next token $v$, when $P_t^T(v)>P_t^S(v)$, the update
can over-support a reference-specific continuation, whereas when
$P_t^S(v)>P_t^T(v)$, it can suppress a valid alternative.
Figure~\ref{fig:teacher-superiority-case} illustrates such a failure at a shared
prefix: the reference-conditioned teacher locally omits a term, while the
outcome-verified student follows a valid alternative.

\FloatBarrier

\subsection{VISTA Training Objective}
\label{sec:method-overview}
A natural mitigation for the problem above is to let the teacher learn from
the student at token positions where the student is closer to the ideal
problem-only distribution, namely where $\Delta_t>0$. However, because of the unobservable nature of the ideal distribution $P_t^I$, $\Delta_t$ cannot be computed during training. As an initial exploration, we introduce
\textbf{V}erifier-\textbf{I}nformed \textbf{S}tudent-to-\textbf{T}eacher
\textbf{A}daptation (\textbf{VISTA}), a simple but effective mechanism that
enables the student to provide training supervision for the teacher.
VISTA uses two simple signals to determine where the teacher update is
allowed: outcome verification identifies rollouts with valid outcomes,
and KL-guided position selection prioritizes token positions with large
teacher--student disagreement. We detail these two selection procedures in
Section~\ref{sec:method-gate}.
Once a trajectory qualifies for feedback, VISTA forms the teacher update from the current student trajectory and paired teacher--student distributions, avoiding a new sampling loop or an independent reward-optimization phase.

\label{sec:method-obj}
To formalize this high-level design as a training objective, we build on the
student and teacher next-token distributions defined in
\S\ref{sec:prelim-opsd}, namely $\PS=\piS(\cdot\mid x,y_{<t})$ and
$\PT=\piT(\cdot\mid x,y^\star,y_{<t})$. VISTA retains the OPSD
student loss from Eq.~\eqref{eq:opsd} and adds a teacher loss over selected
token positions. Let
$\gamma_t(x,y,y^\star)\in\{0,1\}$ denote the binary mask that determines whether
the teacher is updated at token position $t$ of rollout $y$. The two
trajectory-level losses are
\begin{equation}
\begin{aligned}
  \Ls(x,y,y^\star)
  &=\frac{1}{|y|}\sum_{t=1}^{|y|}
    D^{\mathrm{clip}}_{\mathcal V,\tau}\!\big(
      \operatorname{sg}[\PT]\Vert\PS\big),
  \\
  \Lt(x,y,y^\star)
  &=\frac{1}{|y|}\sum_{t=1}^{|y|}\gamma_t(x,y,y^\star)
    D^{\mathrm{clip}}_{\mathcal V,\tau}\!\big(
      \PT\Vert\operatorname{sg}[\PS]\big).
\end{aligned}
\label{eq:trajectory-losses}
\end{equation}
Here, $\operatorname{sg}[\cdot]$ denotes stop-gradient: it blocks gradients
through $\PT$ in $\Ls$ and through $\PS$ in $\Lt$, so the two losses update only
the student and teacher, respectively.
Both losses are evaluated from the same pre-update policy snapshot, with the
sampled rollout and teacher-update mask treated as fixed.
$\Ls$, which places the optimized distribution $\piS$ in the second argument of the KL divergence, is commonly known as the forward-KL objective.
Its mode-covering tendency encourages the
student to retain probability across teacher-supported alternatives. 
By contrast, $\Lt$ optimizes the first argument of the KL divergence, yielding a reverse-KL objective. This design exploits the mode-seeking tendency of reverse-KL, encouraging the teacher to concentrate more of its probability mass on fewer alternatives in the student's next-token distribution~\cite{gu2024minillm}.

Combining the student and teacher losses yields the complete VISTA objective:
\begin{equation}
  \Ltot
  =
  \Ex_{\substack{(x,y^\star)\sim\mathcal{S}\\
                  y\sim\piS(\cdot\mid x)}}
  \left[
    \Ls(x,y,y^\star)+\lambda\Lt(x,y,y^\star)
  \right],
  \label{eq:total}
\end{equation}
where $\lambda$ is a positive weight controlling the contribution of the teacher-adaptation loss.

\subsection{Outcome Gating and KL-Guided Position Selection}
\label{sec:method-gate}
As defined in Sec.~\ref{sec:method-obj}, VISTA incorporates the
teacher-adaptation loss $\Lt$ into its training objective. However, it remains
to determine on which rollouts and at which token positions the teacher should
be updated.
Accordingly, we factorize the binary teacher-update mask $\gamma_t$ for each generated token $y_t$ into rollout-level and token-level components:
\begin{equation}
  \gamma_t(x,y,y^\star)
  =
  \gamma^{\mathrm{out}}(x,y)\,
  \gamma^{\mathrm{pos}}_t(x,y,y^\star).
  \label{eq:mask-factor}
\end{equation}
While the outcome gate $\gamma^{\mathrm{out}}(x,y)$ determines whether
the completed rollout $y$ is eligible to train the teacher at all, the
position mask $\gamma^{\mathrm{pos}}_t(x,y,y^\star)$ determines whether the
teacher loss is calculated at this token position. 

\paragraph{Outcome gate.}
Student supervision is not uniformly reliable: when a rollout fails, the
verifier reveals that the trajectory contains an error but not where it occurs.
Using its token distributions to update the teacher could then reinforce the
unidentified error. VISTA therefore enables teacher adaptation only when a
deterministic verifier accepts the rollout outcome. In mathematics and code,
such verification requires only rule-based answer checks or sandboxed test
execution, leaving no need for a learned reward model or process-level
annotations. Specifically, it sets
$\gamma^{\mathrm{out}}(x,y)=1$ if the verifier accepts the outcome of rollout
$y$, and $0$ otherwise. Passing this gate provides outcome-level evidence that
the student can reach a valid solution from the problem alone.
The student distributions along that rollout can therefore serve as feedback
for adapting the privileged teacher.

\paragraph{Top-$k$ KL-divergence position mask.}
Teacher adaptation must remain selective: updating the teacher at every rollout
token can quickly shrink the teacher--student gap, weakening supervision in
subsequent OPSD steps and potentially hurting the student's final performance.
VISTA therefore evaluates the teacher loss only at rollout positions where
the \textbf{two distributions disagree most}, which it identifies by assigning
each position a teacher--student KL score:
\begin{equation}
  d_t(x,y,y^\star)
  =D_{\mathcal V}\!\big(\PT\Vert\PS\big),
  \label{eq:position-score}
\end{equation}
It then selects up to $k$ highest-scoring positions along the rollout:
\begin{equation}
  \begin{aligned}
  \mathcal I_k(x,y,y^\star)
  &=\operatorname{TopK}_{j\in\{1,\ldots,|y|\}}^{\min(k,|y|)}
    d_j(x,y,y^\star),
  \\
  \gamma^{\mathrm{pos}}_t(x,y,y^\star)
  &=\mathbb{I}\!\left[t\in\mathcal I_k(x,y,y^\star)\right].
  \end{aligned}
  \label{eq:topgap}
\end{equation}
Consequently, $\gamma_t(x,y,y^\star)=1$ only at the selected top-$k$
KL-divergence positions on outcome-verified rollouts, and is $0$
otherwise.

\FloatBarrier
\section{Experiments}
\label{sec:experiments}

We evaluate VISTA on competition-level mathematical reasoning by asking whether
student-to-teacher adaptation improves upon OPSD across model scales and how
much the verifier-based outcome gate and top-$k$ KL-divergence position mask contribute to
these gains. We further conduct three teacher-side analyses, which show that
VISTA's adapted teacher provides stronger token-level support for valid student
reasoning, exhibits explicit reference attribution less frequently, and
preserves its privileged problem-solving capability.

\subsection{Experimental Settings}
\label{sec:setup}

\paragraph{Models and data.}
We evaluate VISTA with the instruct-tuned Qwen3-1.7B, Qwen3-4B, and
Qwen3-8B models~\cite{qwen3}. Following OPSD~\cite{opsd2026}, we train on the
OpenThoughts mathematical-reasoning corpus~\cite{guha2025openthoughts} with the
same 100-step on-policy distillation budget and student-training protocol,
changing only the student learning rate for our setup.

\paragraph{Evaluation.}
We evaluate on AIME 2024, AIME 2025, and HMMT 2025
~\cite{aime2024,aime2025,balunovic2025matharena} and report Avg@12, the mean
accuracy over 12 sampled completions. Table~\ref{tab:main} compares VISTA with
the base model, SFT, GRPO~\cite{shao2024deepseekmath}, and standard OPSD using
results reported in the OPSD paper~\cite{opsd2026}. Additionally, we train and
evaluate SDPO~\cite{hubotter2026sdpo} under the same protocol.

\paragraph{Teacher adaptation.}
For teacher training, we keep most training hyperparameters at the default
values used for the OPSD student, while varying only the teacher-loss weight,
$\lambda$, and the position budget, $k$.
Across the model scales and values tested, we empirically find that settings
with $\lambda\geq0.75$ and $k\in[16,32]$ provide a strong, transferable
operating range for teacher adaptation.

\begin{table*}[t!]
\centering
\small
\setlength{\tabcolsep}{10pt}
\begin{tabular}{llcccc}
\toprule
Model & Method & AIME24 & AIME25 & HMMT25 & Average \\
\midrule
\multirow{6}{*}{Qwen3-1.7B}
 & \quad Base       & 51.5 & 36.7 & 23.1 & 37.1 \\
 & \quad SFT        & 48.4 & 36.3 & 22.7 & 35.8 \\
 & \quad GRPO       & 51.1 & 38.3 & 23.7 & 37.7 \\
 & \quad SDPO       & 52.8 & 36.9 & 22.8 & 37.5 \\
 & \quad OPSD       & \underline{57.2} & \textbf{43.9} & \underline{29.2} & \underline{43.4} \\
 & \quad VISTA      & \textbf{57.8} & \underline{43.3} & \textbf{30.8} & \textbf{44.0} \\
\midrule
\multirow{6}{*}{Qwen3-4B}
 & \quad Base       & 74.9 & 66.4 & 42.2 & 61.2 \\
 & \quad SFT        & 70.2 & 62.3 & 43.4 & 58.6 \\
 & \quad GRPO       & 75.6 & 68.1 & 44.4 & 62.7 \\
 & \quad SDPO       & 76.1 & 65.8 & 42.5 & 61.5 \\
 & \quad OPSD       & \underline{76.4} & \underline{68.3} & \underline{46.1} & \underline{63.6} \\
 & \quad VISTA      & \textbf{77.2} & \textbf{69.2} & \textbf{46.4} & \textbf{64.3} \\
\midrule
\multirow{6}{*}{Qwen3-8B}
 & \quad Base       & 75.8 & 65.6 & 43.9 & 61.8 \\
 & \quad SFT        & 72.3 & 64.2 & 42.9 & 59.8 \\
 & \quad GRPO       & 76.4 & 68.9 & \underline{46.7} & 64.0 \\
 & \quad SDPO       & 76.4 & 68.1 & 44.4 & 63.0 \\
 & \quad OPSD       & \underline{77.8} & \underline{70.8} & 45.8 & \underline{64.8} \\
 & \quad VISTA      & \textbf{78.6} & \textbf{73.9} & \textbf{48.3} & \textbf{66.9} \\
\bottomrule
\end{tabular}
\caption{Main results on competition-level mathematical reasoning (Avg@12,
\%). Each benchmark uses its peak checkpoint under the OPSD protocol, and
Average is the unweighted mean over the three benchmarks. Base, SFT, GRPO, and
OPSD results are from the OPSD paper~\cite{opsd2026}; all post-training methods
use the same corpus, with baseline budgets retained as reported. The best and
second-best results within each model block are shown in boldface and
underlined, respectively.}
\label{tab:main}
\end{table*}

\subsection{Main Results}
\label{sec:results}

Under the matched OPSD protocol, Table~\ref{tab:main} shows that VISTA delivers
the strongest and broadest performance among all compared methods. It achieves
the highest three-benchmark Avg@12 at every Qwen3 scale, improving over reported
OPSD by $0.6$, $0.7$, and $2.1$ points at 1.7B, 4B, and 8B, respectively. Across
the nine scale--benchmark settings, VISTA sets eight new state-of-the-art results.
The advantage is especially pronounced at 8B, where VISTA surpasses reported
OPSD by $0.8$ points on AIME24, $3.1$ on AIME25, and $2.5$ on HMMT25.

The substantially larger gain at 8B highlights VISTA's favorable scaling with
student capacity. As the student becomes stronger, its successful rollouts
provide higher-quality distributions from which the privileged teacher can
learn. The improved teacher, in turn, provides stronger supervision for
subsequent student updates. This positive loop helps account for VISTA's
pronounced advantage at the largest scale.

\subsection{Ablation Studies}
\label{sec:ablation}

To verify the effectiveness of VISTA's outcome-gating and position-selection
designs, we conduct the following ablation studies using the Qwen3-8B model
under the training and evaluation protocol described in \S\ref{sec:setup}.

\paragraph{Outcome gate.}
To separate the semantics of outcome gating from the number of teacher updates,
we compare VISTA with three controls: adapting the teacher on every rollout, on
a random $60\%$ subset that approximately matches VISTA's activation rate, or
only on verifier-rejected rollouts. 

\begin{table}[t]
\centering
\footnotesize
\setlength{\tabcolsep}{3pt}
\begin{tabular}{@{}lcc@{}}
\toprule
Adaptation rule & Rate & Avg@12 $\uparrow$ \\
\midrule
\quad OPSD & $0\%$ & 64.8 \\
\midrule
\quad All rollouts & $100\%$ & 65.8 \\
\quad Random gate & $60\%$ & 65.6 \\
\quad Reverse gate & $\sim40\%$ & 64.5 \\
\quad VISTA & $\sim60\%$ & \textbf{66.9} \\
\bottomrule
\end{tabular}
\caption{Outcome-gate ablation on Qwen3-8B. Rate denotes the fraction of
rollouts routed to teacher adaptation; the random gate approximately matches
VISTA's rate, while the reverse gate uses only verifier-rejected rollouts.}
\label{tab:gate-ablation}
\end{table}

Even without outcome-aware selection, adapting the teacher improves over standard OPSD, as random gating and all-rollout adaptation raise Avg@12 from 64.8 to 65.6 and 65.8, respectively. By moving the teacher distribution closer to the student's, adaptation makes the resulting teacher targets easier for the student to learn.

\begin{figure}[t]
  \centering
  \includegraphics[width=\linewidth]{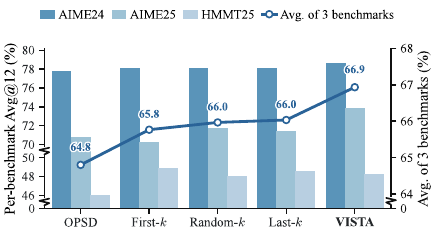}
  \caption{Rollout-position selection ablation on Qwen3-8B (Avg@12). We compare
  OPSD with four teacher-adaptation variants: (A) the first $k$
  positions, (B) $k$ uniformly random positions, (C) the
  last $k$ positions, and (D) the top-$k$ KL-divergence positions.
  VISTA uses (D) and attains the best three-benchmark average.}
  \label{fig:position-ablation}
\end{figure}

We next ask which types of student trajectories are most beneficial for teacher adaptation. The two outcome-directed gates yield opposite results: the reverse gate, which adapts the teacher only on verifier-rejected rollouts, drops to $64.5$, below OPSD, whereas VISTA, which uses outcome-verified rollouts, achieves the best result among all variants at $66.9$ Avg@12. This contrast shows that outcome verification provides a simple yet effective criterion for selecting high-quality student trajectories, along which the student's token distributions offer reliable supervision for teacher adaptation.
\paragraph{Rollout-position selection.}
To identify the positions along an outcome-verified rollout at which the student next-token distributions provide the most useful supervision for the teacher, we compare four rules: selecting the first $k$ positions, $k$ uniformly sampled positions, the last $k$ positions, and VISTA’s KL-guided top-$k$ positions.
Figure~\ref{fig:position-ablation}
reports the per-benchmark results and their average score.

All four position-selection variants outperform OPSD’s 64.8 Avg@12, with VISTA’s KL-guided top-(k) rule achieving the best score of 66.9. 
Large KL divergence identifies the positions of greatest teacher--student disagreement. When these positions occur in an outcome-verified rollout, the valid outcome lends credibility to the student’s reasoning. The disagreement is therefore more likely to expose teacher distributions misaligned with valid problem-only reasoning, making the corresponding student distributions especially useful feedback for the teacher.

\begin{figure}[t]
  \centering
  \includegraphics[width=\linewidth]{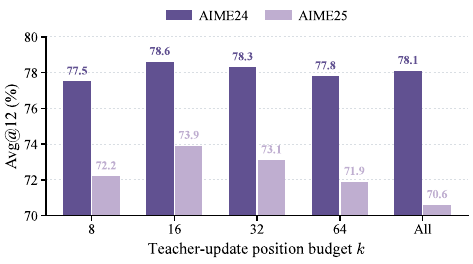}
  \caption{Avg@12 on AIME24 and AIME25 as the number of teacher-update
  positions varies.}
  \label{fig:topk-position-budget}
\end{figure}

\paragraph{Top-$k$ selection size.}
Having established the advantage of KL-guided position selection, we next examine how the number of selected positions affects the effectiveness of teacher adaptation.
We evaluate $k\in\{8,16,32,64\}$ and an $\mathrm{All}$ setting, which applies the teacher loss at every rollout position within the first $1{,}024$ tokens, following OPSD. Figure~\ref{fig:topk-position-budget} summarizes the
resulting Avg@12 scores on AIME24 and AIME25.

VISTA achieves its best performance at $k=16$, with Avg@12 scores of $78.6$ on AIME24 and $73.9$ on AIME25.
The $k=32$ setting performs comparably. 
Across the full sweep, both benchmarks exhibit an overall rise-then-fall trend, as selection sizes below $16$ and above $32$ consistently underperform the two intermediate choices. 

One plausible explanation is that this trend reflects a trade-off in teacher adaptation. 
With too few selected positions, the teacher update remains too
sparse to support sufficient adaptation. By contrast, selecting too many positions pulls the teacher distribution too rapidly toward the student's, which can erode the teacher's distinct supervisory signal and make subsequent student updates less effective.

\subsection{Analytical Studies}
\label{sec:analysis}

Main-task accuracy alone cannot tell us whether adaptation brings $P_t^T$ closer
to the ideal distribution $P_t^I$ defined in \S\ref{sec:feedback-gap}, and the
unobservable nature of $P_t^I$ makes $D_{\mathrm{KL}}(P_t^I\Vert P_t^T)$
impossible to compute during training. We therefore turn to three complementary
analyses of observable teacher behavior for indirect evidence that adaptation
moves $P_t^T$ in the intended direction.

\paragraph{Token-level teacher support for valid student reasoning.}
We construct a fixed diagnostic set by sampling problems from the
training set. 
Using the base Qwen3-8B model in the problem-only student setting,
we generate candidate rollouts and retain 100 trajectories whose final answers
are verified to be correct and whose complete reasoning traces are manually
confirmed to be sound. 
The teacher should therefore not assign low probability
to too many tokens along these trajectories. 
We evaluate the teacher checkpoints
saved throughout the VISTA run using the same hyperparameters and configuration
as in training, thereby mirroring the training setup. At each checkpoint, we
count positions satisfying
$\piT(y_t\mid x,y^\star,y_{<t})\leq 0.05$ and average the count over the 100
trajectories.

\begin{figure}[t]
  \centering
  \includegraphics[width=\linewidth,height=1.35in,keepaspectratio]{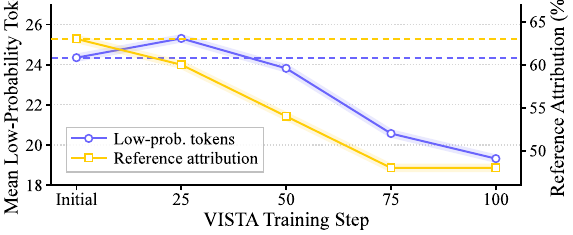}
  \caption{Teacher-adaptation diagnostics across Qwen3-8B checkpoints:
  low-probability tokens ($\piT(y_t)\leq 0.05$) on validated student
  trajectories and explicit reference attribution in teacher responses.}
  \label{fig:teacher-adaptation-diagnostics}
\end{figure}

Figure~\ref{fig:teacher-adaptation-diagnostics} shows that the mean count decreases over the course of training, reaching 19.32 at the final checkpoint, representing a reduction of 5.03 relative to
the base model.
This indicates that adaptation provides increasingly effective token-level support for valid student-generated trajectories throughout training.

\paragraph{Explicit reference attribution during teacher adaptation.}
To track the teacher's undesirable propensity to explicitly invoke the privileged reference, we measure 
the frequency of \emph{explicit reference attribution}, defined as the percentage of sampled responses
in which the teacher explicitly acknowledges the privileged reference. 
The evaluation covers the initial Qwen3-8B teacher and checkpoints saved throughout adaptation on 100 randomly sampled training problems. At each evaluation point, responses are sampled from the teacher for
every problem, conditioned on both the problem and its reference solution.

As shown in Figure~\ref{fig:teacher-adaptation-diagnostics}, the rate of
explicit reference attribution decreases over training and reaches $48\%$ at
both steps 75 and 100, 15 percentage points below the initial teacher's $63\%$.
This decline indicates that VISTA's teacher adaptation weakens its preference for reference-dependent continuations, thereby providing guidance better suited to the student's problem-only distribution.

\paragraph{Teacher capability during training.}
Throughout training, we evaluate the teacher’s performance on the validation set by providing it with reference answers. As shown in Figure 7, the full trajectory maintains a perfect Pass@12 throughout training, while Avg@12 shows a gradual upward trend. 
These results show that adapting the teacher to the student does not compromise the
teacher's own problem-solving performance.

\begin{figure}[htbp]
  \centering
  \includegraphics[width=\linewidth]{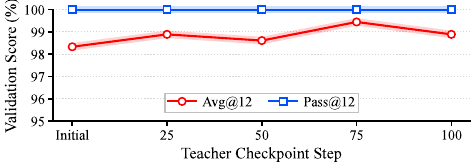}
  \caption{Privileged-teacher validation of the Qwen3-8B teacher during training.}
  \label{fig:teacher-validation}
\end{figure}

\FloatBarrier

\section{Conclusion}
\label{sec:conclusion}

VISTA revises the one-way supervision structure of OPSD by enabling selective student-to-teacher adaptation. Within outcome-verified rollouts, VISTA
adapts the privileged teacher toward the student next-token distributions at the top-$k$ positions with the largest teacher--student KL divergence, while preserving
the standard student update. The method reuses the existing rollout and KL objective without additional sampling or a separate reward objective, and retains only the
problem-only student at inference. Across three Qwen3 scales and three competition-math benchmarks, VISTA achieves the highest Avg@12 at every scale. Ablations show
that outcome-verified, high-disagreement feedback is more effective than indiscriminate or random teacher adaptation. Teacher-side analyses further indicate better support for validated student trajectories, less explicit reference attribution, and preserved privileged-teacher capability. These results suggest that privileged supervision in OPSD need not remain fixed, but can be improved through selective feedback from the student it trains.

\clearpage
\bibliography{references}

\clearpage
\appendix
\section{Use of AI Systems}
\label{app:ai-use}

Generative AI tools were used only for language editing and organization of
this supplementary document.  No AI was used to conceive, propose, or
formulate the VISTA method.  No AI was used to choose experimental settings or
to design and plan experiments.  The authors reviewed every AI-assisted edit and remain fully
responsible for all technical claims, numerical results, citations, and final
submitted wording.

\section{Implementation Details}
\label{app:implementation}

\subsection{Complete VISTA Training Procedure}
\label{app:algorithm}

The main paper gives the VISTA objective and its two selection rules.  This
subsection makes the corresponding minibatch procedure and gradient routing
explicit.

For a training pair $(x,y^\star)$, the problem-only student samples one rollout
$y$.  The student and privileged teacher are then evaluated at the same student
prefixes.  The student loss is active at every rollout position.  The teacher
loss is active only if the deterministic outcome verifier accepts the completed
rollout, and then only at the $\min(k,|y|)$ positions with the largest raw
teacher-first KL scores.

Algorithm~\ref{alg:vista-supp} shows where this procedure enters the training
pipeline.  At every minibatch update, VISTA retains the standard OPSD student
branch over all rollout positions and adds the verifier-gated, top-$k$
teacher-adaptation branch before the two parameter blocks are updated
synchronously.

For convenience, the implementation uses the teacher-rate argument $\eta_T$
instead of $\lambda$ to encode the same relative update strength.  With student
rate $\eta_S$, the main-paper coefficient is simply
$\lambda=\eta_T/\eta_S$.

\begin{algorithm}[t]
\caption{VISTA minibatch training procedure.}
\label{alg:vista-supp}
\footnotesize
\begin{algorithmic}[1]
\raggedright
\REQUIRE Training set $\mathcal S$; separately parameterized student
$(\piS,\theta_S)$ and teacher $(\piT,\theta_T)$; verifier $V$; update budget
$M$; position budget $k$; clipping threshold $\tau$; student rate $\eta_S$;
teacher-rate argument $\eta_T$
\FOR{$m=1,\ldots,M$}
  \STATE Sample minibatch $\mathcal B\subset\mathcal S$ and store the
  pre-update snapshot $(\theta_S^m,\theta_T^m)$
  \STATE Set $\widehat{\Ls}\gets0$, $\widehat{\Lt}\gets0$, and $n_T\gets0$
  \FORALL{$(x,y^\star)\in\mathcal B$}
    \STATE Sample $y\sim\piS(\cdot\mid x)$; compute $\gamma^{\mathrm{out}}\gets V(x,y)$;
    evaluate $P_t^S$ and $P_t^T$ for all $t$
    \STATE Add $|y|^{-1}\sum_tD^{\mathrm{clip}}_{\mathcal V,\tau}
    (\operatorname{sg}[P_t^T]\Vert P_t^S)$ to $\widehat{\Ls}$
    \IF{$\gamma^{\mathrm{out}}=1$}
      \STATE Rank positions by $d_t=D_{\mathrm{KL}}(P_t^T\Vert P_t^S)$ and
      let $\mathcal I_k$ contain the largest $\min(k,|y|)$ scores
      \STATE Add $\sum_{t\in\mathcal I_k}D^{\mathrm{clip}}_{\mathcal V,\tau}
      (P_t^T\Vert\operatorname{sg}[P_t^S])$ to $\widehat{\Lt}$ and add
      $|\mathcal I_k|$ to $n_T$
    \ENDIF
  \ENDFOR
  \STATE Compute $g_S\gets\nabla_{\theta_S}\widehat{\Ls}$ and, if $n_T>0$,
  $g_T\gets\nabla_{\theta_T}\widehat{\Lt}$
  \STATE Update $\theta_S$ using $g_S$ and $\eta_S$; if $n_T>0$, update
  $\theta_T$ using $g_T$ and $\eta_T$
\ENDFOR
\ENSURE Trained problem-only student parameters $\theta_S$
\end{algorithmic}
\end{algorithm}

\subsection{Training Configuration}
\label{app:hyperparameters}

\paragraph{OPSD lineage and VISTA modification.}
Our implementation retains the central OPSD training logic and student-side
pipeline~\cite{opsd2026}.  The key difference from standard OPSD lies in the
teacher branch: whereas OPSD keeps the privileged target fixed, VISTA maintains
an independent, trainable teacher and adds the verifier-gated, top-$k$ teacher
update from Section~\ref{app:algorithm}.  Apart from this teacher-side addition,
the student branch follows OPSD and is optimized at every valid rollout position.

\paragraph{Data, conditioning, and rollouts.}
Student and teacher are initialized from the same instruct-tuned Qwen3 model at
the 1.7B, 4B, or 8B scale~\cite{qwen3}.  We draw from a pool of up to 30K
problem--reference-solution pairs from the mathematical-reasoning subset of
OpenThoughts~\cite{guha2025openthoughts}.  With one rollout per prompt, an
effective global batch size of 32, and 100 updates, the reported schedule
samples about $3{,}200$ prompt--rollouts in total.  The student receives only
the problem, with thinking mode disabled.  The teacher receives the problem and
reference solution, with thinking mode enabled.  Verification uses the full
response (up to 4,096 new tokens), whereas next-token loss is computed only on
its first 1,024 tokens: the student scores these tokens conditioned on the
problem-only prompt, and the teacher scores the same tokens conditioned on the
problem and reference solution.  Table~\ref{tab:training-configuration} provides
a more detailed listing of the implementation hyperparameters and related
settings.

\paragraph{Prompt templates.}
We next detail how the student and teacher prompts are constructed during
training.  We use exactly the same prompt-construction procedure as
OPSD~\cite{opsd2026}.  In the default launcher path,
\texttt{reason\_first=False}.  The literal message contents are shown below; the
student uses thinking mode off, while the teacher uses thinking mode on.

\begin{tcolorbox}[vistaPromptBox,title={Student training prompt (thinking disabled)}]
\texttt{Problem: \{problem\}}\\[-1pt]
\texttt{Please reason step by step, and put your final answer}\\[-1pt]
\texttt{within \textbackslash boxed\{\}.}
\end{tcolorbox}

\begin{tcolorbox}[vistaPromptBox,title={Teacher training prompt (thinking enabled)}]
\texttt{Problem: \{problem\}}\\[2pt]
\texttt{=== Reference Solution Begin ===}\\[-1pt]
\texttt{\{reference\_solution\}}\\[-1pt]
\texttt{=== Reference Solution End ===}\\[2pt]
After reading the reference solution above, make sure you truly understand the
reasoning behind each step -- do not copy or paraphrase it.  Now, using your own
words and independent reasoning, derive the same final answer to the problem
above.  Think step by step, explore different approaches, and do not be afraid
to backtrack or reconsider if something does not work out.\\[2pt]
\texttt{Please reason step by step, and put your final answer}\\[-1pt]
\texttt{within \textbackslash boxed\{\}.}
\end{tcolorbox}

\paragraph{Compute environment and training cost.}
The Qwen3-1.7B, Qwen3-4B, and Qwen3-8B post-training runs were conducted on one
node with eight NVIDIA H800 80GB GPUs.  Complete 100-update training runs from Qwen3-1.7B
through Qwen3-8B took approximately 1--6 hours, depending on model scale.  The
training software stack comprised Python 3.12, CUDA 12.8, PyTorch 2.8,
\texttt{verl} 0.6.1, and vLLM 0.11.0.

\begin{table}[t]
\centering
\footnotesize
\setlength{\tabcolsep}{3pt}
\renewcommand{\arraystretch}{1.02}
\begin{tabular}{@{}p{0.27\columnwidth}p{0.68\columnwidth}@{}}
\toprule
Parameter & Setting \\
\midrule
Model scale & Qwen3-1.7B, Qwen3-4B, or Qwen3-8B. \\
Data pool & OpenThoughts mathematical-reasoning pool. \\
Student prompt & Problem only; thinking off. \\
Teacher prompt & Problem plus reference solution; thinking on. \\
Temperature & $1.1$. \\
Top-$p$ & $0.95$. \\
Top-$k$ & $20$. \\
Verification rollout & Up to $4{,}096$ new tokens for correctness checking. \\
Scored rollout tokens & Up to the first $1{,}024$ tokens of each rollout. \\
Distillation & Full-vocabulary forward KL with $\beta=0$. \\
Clip threshold & $\tau=0.05$. \\
Teacher gate & Verifier-accepted rollouts only. \\
Position selection & Raw-KL top-$k$ positions; $k=32$ for 1.7B/4B and $k=16$
for 8B. \\
Learning rates & $\eta_S=4\times10^{-6}$ and $\eta_T=3\times10^{-6}$;
$\lambda=\eta_T/\eta_S=0.75$. \\
LoRA & $r=64$, $\alpha=128$; targets
\texttt{q\_proj}, \texttt{k\_proj}, \texttt{v\_proj}, \texttt{o\_proj},
\texttt{gate\_proj}, \texttt{up\_proj}, \texttt{down\_proj}. \\
Adapters & Independent student and teacher adapters. \\
Optimizer & AdamW; zero weight decay; linear student schedule without warmup. \\
Budget & Effective global batch $32$; 100 updates. \\
Numerics & Bfloat16 and gradient checkpointing. \\
Hardware & One node with eight H800 80GB GPUs for every model scale. \\
Checkpoints & Saved every 25 updates. \\
\bottomrule
\end{tabular}
\caption{Core VISTA training configuration.}
\label{tab:training-configuration}
\end{table}

\subsection{Evaluation Configuration}
\label{app:evaluation-configuration}

\paragraph{OPSD evaluation lineage.}
Our implementation directly reuses the evaluation logic released with
OPSD~\cite{opsd2026}, including its problem prompt, Qwen3 chat templating,
boxed-answer extraction, mathematical-equivalence checker, and Avg@$N$/Pass@$N$
aggregation.

\paragraph{Benchmarks, prompts, and decoding.}
We evaluate all 30 problems from each of AIME 2024, AIME 2025, and HMMT
2025~\cite{aime2024,aime2025,balunovic2025matharena}.  Each user prompt contains
the problem followed by the following fixed instruction:
\begin{tcolorbox}[vistaPromptBox,title={Evaluation instruction (thinking enabled)}]
\texttt{Please reason step by step, and put your final answer}\\[-1pt]
\texttt{within \textbackslash boxed\{\}.}
\end{tcolorbox}
For every problem, a single vLLM sampling call draws 12 completions
with temperature $1.0$, top-$p=0.95$, top-$k$ disabled, min-$p=0$, and seed
42.  Each completion may contain at most 38,912 new tokens.

\paragraph{Answer checking and metrics.}
To score a completion, the reused OPSD evaluator extracts its final balanced
\texttt{\textbackslash boxed\{\ldots\}} expression and uses
\texttt{math\_verify} to compare the parsed answer with the ground truth for
mathematical equivalence.  A completion with no such expression is counted as
incorrect.
For benchmark $b$ with $N_b=30$ problems and $R=12$ completions per problem,
we report
\begin{align}
\operatorname{Avg@12}(b)
&=\frac{100}{N_bR}\sum_{i=1}^{N_b}\sum_{r=1}^{R}
\mathbb I[\text{completion}_{i,r}\text{ is correct}],
\label{eq:avg12}\\
\operatorname{Pass@12}(b)
&=\frac{100}{N_b}\sum_{i=1}^{N_b}
\mathbb I[\exists r:\ \text{completion}_{i,r}\text{ is correct}].
\label{eq:pass12}
\end{align}

\begin{table}[t]
\centering
\small
\setlength{\tabcolsep}{4pt}
\renewcommand{\arraystretch}{1.03}
\begin{tabular}{@{}ll@{}}
\toprule
\textbf{Parameter} & \textbf{Value} \\
\midrule
Max new tokens & 38,912 \\
Thinking mode & Enabled \\
Temperature & 1.0 \\
Top-$p$ & 0.95 \\
Top-$k$ & $-1$ (disabled) \\
Min-$p$ & 0.0 \\
Seed & 42 \\
Samples per prompt & 12 \\
\bottomrule
\end{tabular}
\caption{VISTA evaluation configuration.}
\label{tab:evaluation-configuration}
\end{table}

\paragraph{Checkpoint reporting and baseline provenance.}
The evaluation phase starts at step 25 and evaluates every saved student
checkpoint whose update is a multiple of 25.  For the 100-update runs reported
in this paper, the evaluated states are therefore steps 25, 50, 75, and 100.
All methods use the same evaluation protocol and peak-selection rule.  Following
the OPSD reporting convention, each benchmark reports its highest Avg@12 over
these states; the main table's \emph{Average} is the unweighted mean of the
three peaks and may combine different steps.  OPSD supplies the Base,
SFT, GRPO, and OPSD entries~\cite{opsd2026}; SDPO and VISTA are evaluated
locally.

\FloatBarrier

\section{Additional Experimental Details}
\label{app:experimental-details}

\subsection{Controlled Ablation Protocols}
\label{app:ablation-protocols}

All controlled ablations use Qwen3-8B and the protocols in
Sections~\ref{app:hyperparameters} and~\ref{app:evaluation-configuration}.

\begin{figure}[t]
  \centering
  \includegraphics[width=\columnwidth]{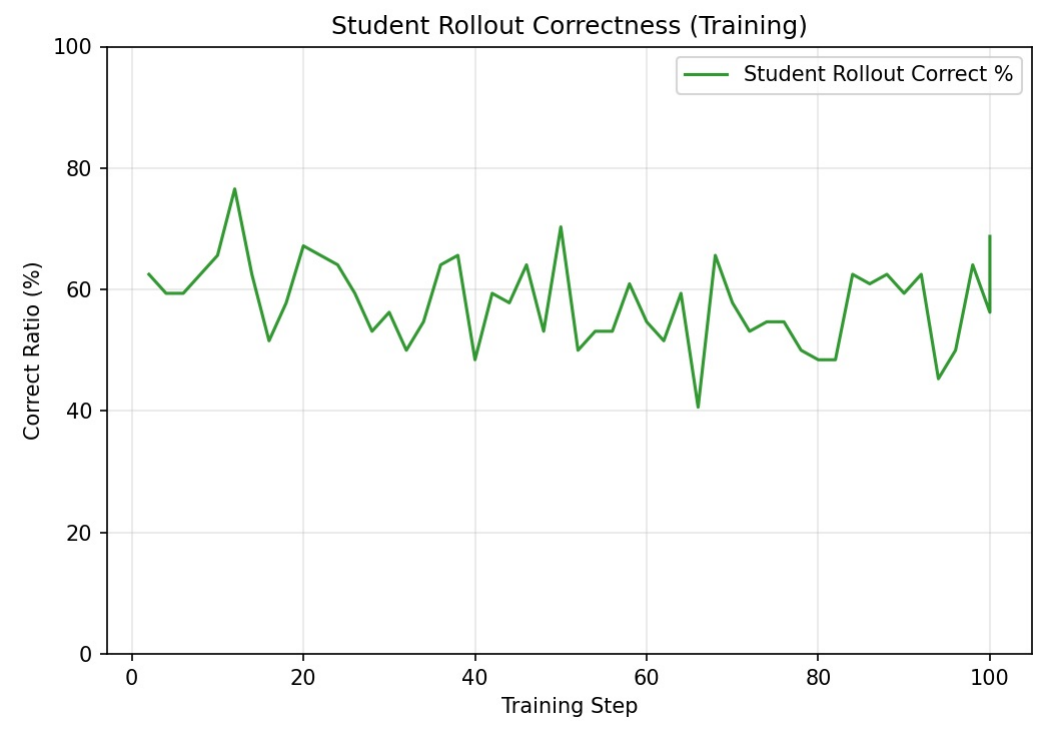}
  \caption{Verifier acceptance rate of Qwen3-8B student rollouts during VISTA
  training, centered near $60\%$.}
  \label{fig:training-rollout-correctness}
\end{figure}

\paragraph{Outcome gate.}
OPSD applies no teacher update.  For the three teacher-adaptation controls and
VISTA, let $V(x,y)\in\{0,1\}$ denote the deterministic outcome verifier,
with $V(x,y)=1$ when the student rollout $y$ for problem $x$ is accepted and
$V(x,y)=0$ otherwise.  The all-rollout control therefore enables the teacher
update on every rollout, whereas the random control independently samples
$r\sim\operatorname{Bernoulli}(0.6)$ for each rollout and enables it when
$r=1$.  The reverse control enables the update only on verifier-rejected
rollouts, while VISTA enables it only on verifier-accepted rollouts.  Equivalently,
the rollout-level masks are
\begin{equation}
\gamma^{\mathrm{out}}(x,y)=\begin{cases}
1, & \text{all-rollout control},\\
r, & \text{random control},\\
1-V(x,y), & \text{reverse control},\\
V(x,y), & \text{VISTA}.
\end{cases}
\label{eq:outcome-gate-controls}
\end{equation}
We use probability $0.6$ for the random control because the verifier accepts
roughly $60\%$ of student rollouts during VISTA training, as shown in
Figure~\ref{fig:training-rollout-correctness}.

\paragraph{Position selection.}
With the gate fixed to verifier-accepted rollouts and $m=\min(k,|y|)$, the
first-$k$ and last-$k$ controls use the first or last $m$ positions, random-$k$
samples an $m$-element subset uniformly without replacement, and the KL control
uses the $m$ largest raw teacher-first KL scores.  More explicitly, for each
rollout position $t$, define
\begin{equation}
d_t \;=\; D_{\mathrm{KL}}\!\left(P_t^T\,\middle\|\,P_t^S\right)
\;=\; \sum_{v\in\mathcal V}P_t^T(v)\log\frac{P_t^T(v)}{P_t^S(v)},
\label{eq:supp-teacher-first-kl}
\end{equation}
where $P_t^T$ and $P_t^S$ are the teacher and student next-token
distributions at the same student-visited prefix.  The controls select
positions as follows: the first-$k$ control uses the first
$m$ rollout positions, the last-$k$ control uses the last $m$ positions, and
the random-$k$ control samples $m$ positions uniformly without replacement.
The KL control ranks all rollout positions by $d_t$ and uses the $m$ positions
with the largest scores.
For this Qwen3-8B position-selection comparison, we fix the position budget to
$k=16$ for every variant.  We do not include a bottom-$k$ control because
its near-zero-KL positions have very small teacher-update losses and provide
little training signal.

\paragraph{Position budget.}
The sweep evaluates $k\in\{8,16,32,64\}$ and an all-position setting.  For
each sparse setting, candidate positions are ranked by the raw teacher-first KL
scores in Eq.~\eqref{eq:supp-teacher-first-kl} and the top $k$ positions are
retained.  The all-position
setting retains every valid position.  For every setting, candidate positions
are restricted to the first $1{,}024$ rollout
tokens.  This limit is
not a separately tuned VISTA choice: it exactly follows the OPSD training
protocol, which computes its distillation loss over the first $1{,}024$ rollout
tokens~\cite{opsd2026}.  The all-position setting applies the teacher loss at
every valid position in that same prefix, so comparisons with OPSD and the
sparse variants use an identical token horizon.
Only AIME 2024 and AIME 2025 were recorded for this sweep.

\FloatBarrier

\subsection{Teacher-Side Analysis Protocols}
\label{app:diagnostic-protocols}

We evaluate the initial Qwen3-8B teacher and the teacher checkpoints saved at
steps 25, 50, 75, and 100 from the specific Qwen3-8B training run whose student
result is reported as the SOTA result in the main table.

\paragraph{Valid student trajectories.}
The fixed bank contains 100 problem-only Qwen3-8B trajectories from randomly
sampled training problems.  Each trajectory is automatically verified as
correct and then manually reviewed; the review confirms that the overwhelming
majority of its tokens and the resulting reasoning trajectory are coherent,
well-formed, and valid for the stated problem.  Using exactly the same
configuration as in training, we then provide each checkpoint teacher with the
paired reference solution and evaluate the token-level probabilities along
these same 100 student trajectories.  We count the tokens assigned probability
at most $0.05$ and average the count over the bank.

\paragraph{Explicit reference attribution.}
At each checkpoint, the privileged teacher samples one reference-conditioned
response for each of 100 randomly sampled training problems.  We use an LLM-as-
judge procedure with Opus 4.6 as the judge model to identify responses whose
realized text explicitly attributes its reasoning to the privileged reference,
including phrases such as ``according to the reference'' or ``as the reference
solution ...''.  We count only trajectories containing such clear reference-
attribution statements.

\paragraph{Privileged-teacher capability.}
The capability probe uses 30 validation problems randomly sampled from the
training set and 12 privileged-teacher responses per problem.

\section{Additional Results}
\label{app:additional-results}

\subsection{Checkpoint-Wise Avg@12 for Qwen3-8B}
\label{app:checkpoint-wise-results}

To characterize how VISTA's model performance evolves over time, we present
the Qwen3-8B student's Avg@12 trajectory on all three benchmarks throughout
training.  Table~\ref{tab:checkpoint-avg12-8b} reports the recorded scores at
steps 25, 50, 75, and 100, allowing us to examine benchmark-specific trends
and the stability of the overall training process.  The scores fluctuate only
modestly, with no abrupt collapse: HMMT25 improves steadily, while AIME24 and
AIME25 remain within comparable ranges.  Overall, VISTA maintains a stable
performance trajectory over training.
\begin{table}[H]
\centering
\small
\setlength{\tabcolsep}{6pt}
\begin{tabular}{lccc}
\toprule
Step & AIME24 & AIME25 & HMMT25 \\
\midrule
25  & 74.4 & \textbf{71.9} & 43.1 \\
50  & 76.9 & 69.4 & 43.9 \\
75  & \textbf{78.6} & \textbf{71.9} & 45.6 \\
100 & 76.4 & 71.1 & \textbf{48.3} \\
\bottomrule
\end{tabular}
\caption{Qwen3-8B VISTA student Avg@12 (\%) by checkpoint.}
\label{tab:checkpoint-avg12-8b}
\end{table}

\FloatBarrier

\subsection{Student Adaptation Scale Under Sparse and All-Position Updates}
\label{app:student-weight-drift}

The purpose of this experiment is to show the phenomenon described in the
main paper: updating the teacher at all positions, or at too many positions,
can make the subsequent student update insufficient.  We compare the
Qwen3-8B top-$16$ and all-position runs at the checkpoints
$t\in\{25,50,75,100\}$.  Let $q\in\{16,\mathrm{all}\}$ denote the teacher
position budget, $R\in\{S,T\}$ the student or teacher, and $m\in\mathcal M$
an adapted LoRA module.  The effective LoRA update relative to the frozen base
weight is
\begin{equation}
  \begin{aligned}
    \Delta W_{R,q,m}(t)
    &=\frac{\alpha}{r}B_{R,q,m}(t)A_{R,q,m}(t)\\
    &=2B_{R,q,m}(t)A_{R,q,m}(t).
  \end{aligned}
  \label{eq:effective-lora-drift}
\end{equation}
with rank $r=64$ and scaling $\alpha=128$.  Since LoRA initializes
$B(0)=0$, these are changes relative to the common base model, not distances
from the preceding checkpoint.

Let $N$ be the total number of entries across all adapted modules.  We report
three base-referenced scale statistics:
\begin{align}
  D^{(2)}_{R,q}(t)
  &:=\left(\sum_{m\in\mathcal M}
      \|\Delta W_{R,q,m}(t)\|_F^2\right)^{1/2},
      \label{eq:lora-global-l2}\\
  D^{(\mathrm{mean})}_{R,q}(t)
  &:=\frac{1}{N}\sum_{m\in\mathcal M}
      \|\Delta W_{R,q,m}(t)\|_1,
      \label{eq:lora-global-mean}\\
  D^{(\max)}_{R,q}(t)
  &:=\max_{m\in\mathcal M}
      \|\Delta W_{R,q,m}(t)\|_{\max}.
      \label{eq:lora-global-max}
\end{align}
where $\|\cdot\|_F$ is the Frobenius norm, $\|\cdot\|_1$ sums entrywise
absolute values, and $\|\cdot\|_{\max}$ selects the largest entrywise
absolute value.  We also report the student-to-teacher ratio
\begin{equation}
  \rho_q(t):=
  \frac{D^{(2)}_{S,q}(t)}{D^{(2)}_{T,q}(t)}.
  \label{eq:lora-student-teacher-ratio}
\end{equation}
Thus, $D^{(2)}$ is the global effective-update norm, $D^{(\mathrm{mean})}$ the
mean absolute entry, $D^{(\max)}$ the largest absolute entry, and $\rho_q$ the
relative student/teacher scale.

\begin{figure}[H]
  \centering
  \includegraphics[width=\linewidth]{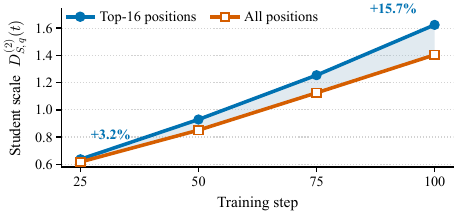}
  \caption{Student effective-weight $\ell_2$ scale,
  $D^{(2)}_{S,q}(t)$.}
  \label{fig:student-l2-drift}
\end{figure}

\begin{figure}[H]
  \centering
  \includegraphics[width=\linewidth]{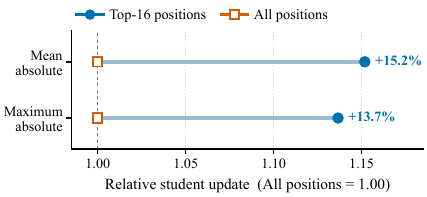}
  \caption{Step-100 ratio of top-$16$ to all-position student effective-weight
  entry scales, $D^{(u)}_{S,16}(100)/D^{(u)}_{S,\mathrm{all}}(100)$, for
  $u\in\{\mathrm{mean},\max\}$.}
  \label{fig:student-entry-scale}
\end{figure}

\begin{figure}[H]
  \centering
  \includegraphics[width=\linewidth]{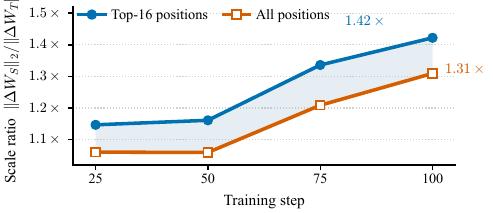}
  \caption{Student-to-teacher effective-weight scale ratio, $\rho_q(t)$.}
  \label{fig:student-scale-ratio}
\end{figure}

\paragraph{Results and interpretation.}
Figure~\ref{fig:student-l2-drift} shows that the top-$16$ student has larger
$D^{(2)}$ at every checkpoint, with its advantage growing from $3.2\%$ at step
25 ($0.637$ versus $0.617$) to $15.7\%$ at step 100 ($1.624$ versus $1.404$).
This widening gap shows that the all-position student accumulates progressively
less adaptation as training proceeds.  The same pattern appears in the step-100
mean and maximum entry scales, which are respectively $15.2\%$ and $13.7\%$
larger for top-$16$ (Figure~\ref{fig:student-entry-scale}), ruling out an effect
driven only by the global norm or a few extreme entries.

The student-to-teacher ratio is also consistently larger for top-$16$
(Figure~\ref{fig:student-scale-ratio}); at step 100, $\rho_{16}=1.42\times$
versus $\rho_{\mathrm{all}}=1.31\times$.  Thus, relative to how far the teacher
has moved, all-position updating leaves the student with less accumulated
change.  These diagnostics support the claim that all-position teacher
adaptation can make subsequent student updates insufficient.

\FloatBarrier

\subsection{Two-Sided Failure of Teacher Superiority}
\label{app:qualitative-case}

\begin{figure*}[p]
  \centering
  \includegraphics[width=\textwidth]{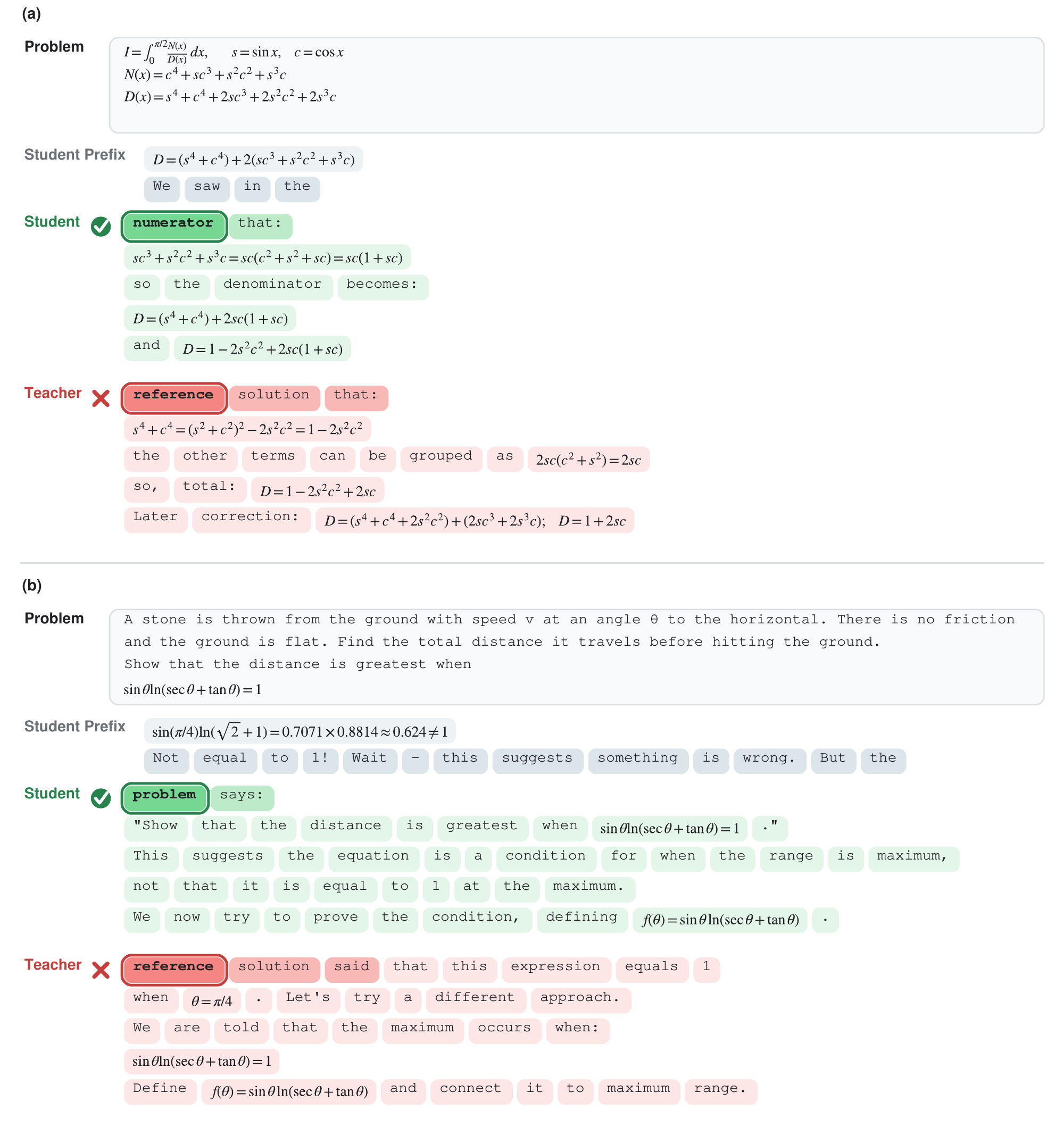}
  \caption{Reference-conditioned over-support: privileged-information leakage
  at a shared student prefix.  At such a prefix, the privileged teacher may
  favor a continuation that depends on information unavailable to the
  problem-only student.}
  \label{fig:teacher-reference-leakage}
\end{figure*}

\begin{figure*}[p]
  \centering
  \includegraphics[width=\textwidth]{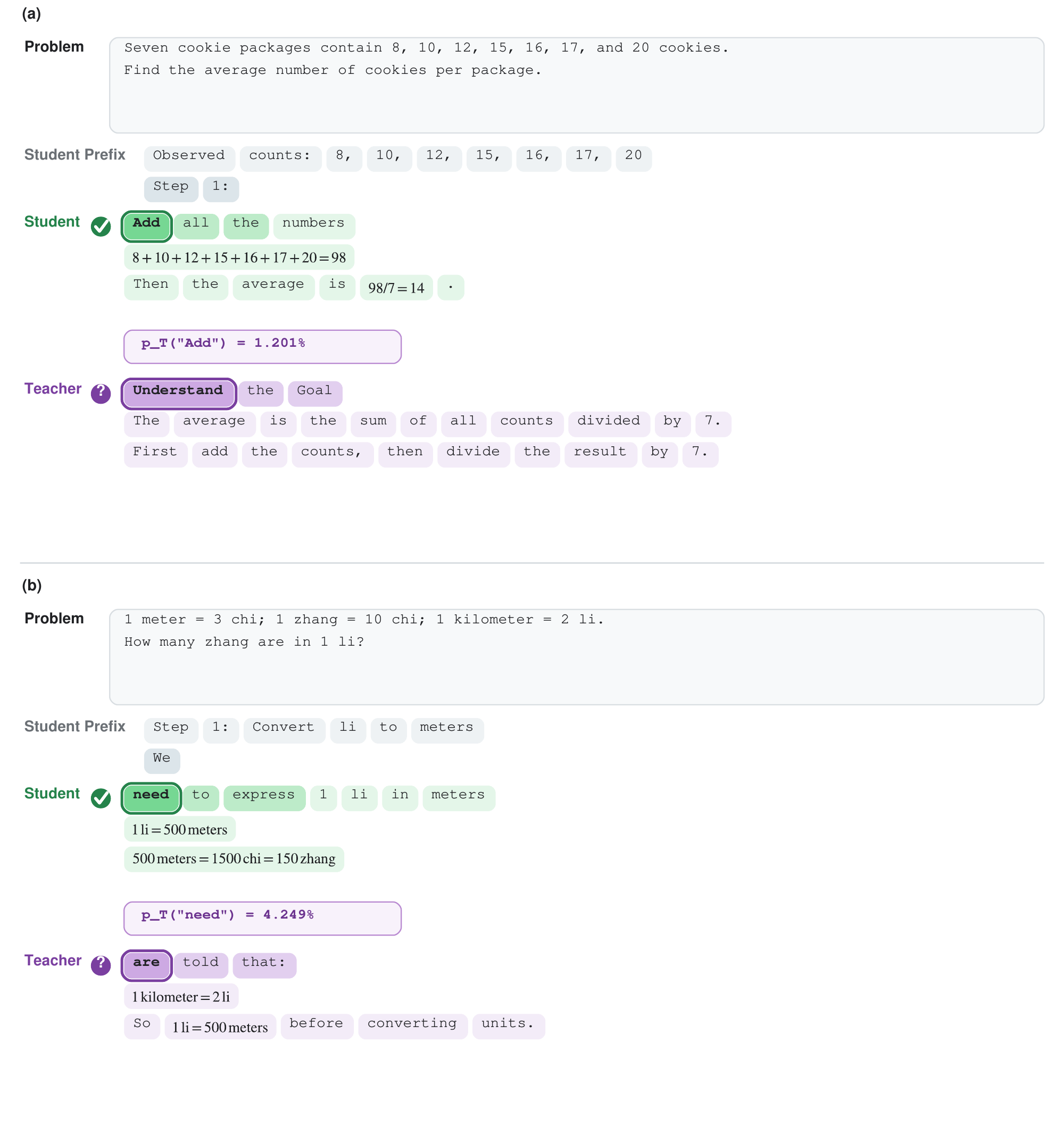}
  \caption{Reference-conditioned under-support: suppression of valid
  problem-only alternatives at a shared student prefix.  At such a prefix,
  the privileged teacher may assign low probability to a locally coherent
  continuation available to the problem-only student.}
  \label{fig:teacher-veto-cases}
\end{figure*}

At each prefix along a student-generated rollout, standard OPSD trains the
student to match the teacher's next-token distribution.  This one-way update
is beneficial only when the teacher's next-token distribution remains a better
learning target than the student's current distribution at every step.
Reference conditioning alone cannot guarantee this:
it can over-support a reference-specific continuation or suppress a valid
problem-only alternative.  Figures~\ref{fig:teacher-reference-leakage} and
\ref{fig:teacher-veto-cases} showcase these two failure modes, respectively,
using the actual Qwen3-8B training setup and examples drawn from the actual
training set.  Each prefix is generated by the problem-only student; the student
and teacher then separately prefill the same prefix and continue generating under
their respective contexts.

The mismatch is two-sided.  In Figure~\ref{fig:teacher-reference-leakage}, the
teacher over-supports a continuation that explicitly invokes the privileged
reference, while the student continues from information available in the
problem and its own derivation.  A one-way student update can then transfer a
reference-dependent shortcut into the problem-only policy.  In the showcased
cases, this dependence appears literally as the teacher opening with ``reference
solution'' or ``reference solution said''.  The problem-only student nevertheless
follows a valid problem-grounded route and reaches the correct answer without
access to the reference.

In Figure~\ref{fig:teacher-veto-cases}, the direction reverses.  The student
offers a locally coherent problem-only next step, but the teacher assigns it
very little probability and favors another continuation.  The same
forward-KL update can therefore veto a viable alternative, reducing the
student's support for a reasoning path that is available at inference time.
In the two showcased cases, the student begins a correct solution with ``Add''
or ``need,'' and both problem-only continuations reach the correct answer.  The
teacher instead opens with ``Understand'' or ``are'' and assigns low probability
to the student's first token.

These examples make concrete why teacher capability or extra information does
not guarantee teacher superiority at every token position, motivating
selective teacher adaptation.

\FloatBarrier

\end{document}